# Parallel Distributed Logistic Regression for Vertical Federated Learning without Third-Party Coordinator

**Shengwen Yang, Bing Ren, Xuhui Zhou, Liping Liu**
Baidu, Inc.
{yangshengwen, renbing, zhouxuhui, liuliping}@baidu.com

## Abstract

Federated Learning is a new distributed learning mechanism which allows model training on a large corpus of decentralized data owned by different data providers, without sharing or leakage of raw data. According to the characteristics of data distribution, it could be usually classified into three categories: horizontal federated learning, vertical federated learning, and federated transfer learning. In this paper we present a solution for parallel distributed logistic regression for vertical federated learning. As compared with existing works, the role of third-party coordinator is removed in our proposed solution. The system is built on the parameter server architecture and aims to speed up the model training via utilizing a cluster of servers in case of large volume of training data. We also evaluate the performance of the parallel distributed model training and the experimental results show the great scalability of the system.

## 1 Introduction

Due to its ability to extract knowledge and insights from massive data and transform data into business values, machine learning has been playing a very important role in various applications, including web search, online advertisement, commodity recommendation, mechanical failure prediction, insurance pricing, and financial risk management, etc. Traditionally, a machine learning model is trained over a centralized corpus of data, which might be collected by a single or multiple data providers. Although parallel distributed algorithms have been developed to accelerate the training process, the training data themselves are still collected and stored centrally in a data center. This way works well when the data are owned by a single data provider or when the data owned by multiple different data providers can be shared and merged together without violating laws and regulations.

In some circumstances, there are some practical challenges to share and circulate data across different organizations. On the one hand, there exists risks of data abuse and/or secondary distribution of data once the data are shared to third parties. On the other hand, some kinds of data are forbidden to be circulated according to certain laws and regulations. How to train machine learning models under these constraints raises a great challenge for both academic and industry communities. Federated learning provides a possible solution for tackling this challenge [McMahan and Ramage, 2017; Yang et al., 2019].

A few years ago, Google proposed a framework for federated learning which aims to enable model training on a large corpus of decentralized data residing on devices like smart phones which allows to both preserve data privacy and prevent data leakage [Konečný et al., 2016]]. The concept of federated learning is extended by Yang et al. [2019] to cover more scenarios and forms a comprehensive secure federated learning framework, including horizontal federated learning (HFL), vertical federated learning (VFL), and federated transfer learning (FTL). HFL is also named sample-based federated learning, which is applicable to the scenarios that datasets owned by different parties share the same feature space but differ in samples. VFL is also named feature-based federated learning, which is suited to the scenarios that datasets owned by different parties share the same sample ID space but differ in feature space. FTL is applicable to the scenarios that datasets owned by different parties differ not only in samples but also in feature space.

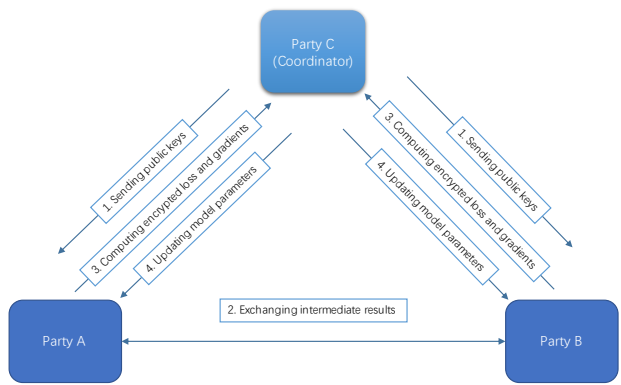

Figure 1: Vertical federated learning with coordinator

In this paper, we introduce a parallel distributed logistic regression implementation for vertical federated learning.

Due to its simplicity and wide usage in many binary classification tasks, there already exist several implementations of logistic regression for vertical federated learning [Hardy et al., 2017]. However, as Figure 1 shows, these existing implementations are built on an architect that involves three parties: party A that has both part of features and labels of samples, party B that has only part of features, and party C that plays as a coordinator, which is very critical role and should be trusted by both party A and B. In our implementation, we propose to remove the party C, i.e. the coordinator, from the system, as shown in Figure 2.

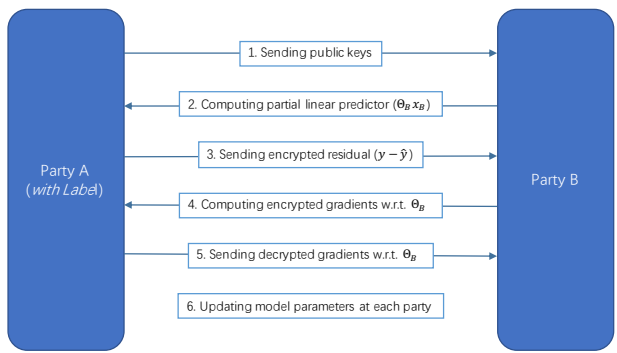

Figure 2: Vertical federated learning without coordinator

The main reasons behind this change are two folds. First, it's very difficult itself to find an authoritative third party that can be trusted by any two parties. Second, it increases the risk of data leakage by involving a third party besides the party A and B. By removing the party C, it can greatly reduce the complexity of the system and decrease the cost of building a joint model between any two parties. We will further discuss the security impact caused by this architectural change. Moreover, in our implementation, we consider to build a parallel distributed system to speed up the training process in the case of very large volume of training data, instead of a system that runs in a single machine at each party.

The main contributions of this paper can be summarized as follows:
1) We propose a new architecture for vertical federated learning that removes the role of third-party coordinator, which greatly reduces the complexity of the system and allows any two parties to train a joint model without the help of a trusted coordinator.
2) We implement a parallel distributed logistic regression for vertical federated learning based on the above architecture, which can handle very large volume of training data by running the training algorithm on a cluster of machines at each party.
3) Besides the two-party model training, our proposed system architecture can also be easily extended to support multi-party model training.

The rest of this paper is organized as follows. In Section 2, we give a brief overview of related works. We present the details of our solution in Section 3. Experimental results and some analysis are given in Section 4. We conclude the paper in Section 5.

## 2 Related Work

This paper is related to federated learning (FL), a new machine learning mechanism that enables to train a joint model on a large corpus of decentralized data owned by different parties while preserving the data privacy. McMahan and Ramage [2017] described the application of FL in the scenario of mobile devices, where the data reside on tens of millions of smart phones and the objective is to train a global model collaboratively by these devices on their local data. Konečný et al. [2016] studied the federated optimization problem for this setting where communication efficiency is of the utmost importance and minimizing the number of rounds of communication is the principal goal. Bonawitz et al. [2019] further presented a high-level design for scalable FL production system in the setting of mobile devices and discussed some challenges, solutions, and open problems in the system design.

Yang et al. [2019] extended the concept of FL and categorized it into three classes according to the pattern of data distribution, including horizontal federated learning (HFL), vertical federated learning (VFL), and federated transfer learning (FTL). It also discussed more application domains of FL such as retail, finance, and healthcare, etc.

This paper is also related to homomorphic encryption (HE) [Rivest et al., 1978; Paillier, 1999; Acar et al., 2018], a cryptographic method which is widely used in the privacy-preserving machine learning algorithms. HE allows computation on encrypted data and the result of such a computation remains encrypted. When decrypted, it matches the result of the computation performed on the plaintext. There are several different types of homomorphic encryption, e.g. fully homomorphic, and partially homomorphic. Among others, additively homomorphic encryption (AHE) is the one which is most commonly used due to its relatively superior computation efficiency. Let the encryption of a number $u$ be $[\![u]\!]$, given any numbers $u$, $v$, and $n$ in plaintext, we have:

$$[\![u]\!] + [\![v]\!] = [\![u + v]\!] \quad (1)$$

$$n \cdot [\![u]\!] = [\![n \cdot u]\!] \quad (2)$$

In some recent works, raw data from different parties are encrypted with AHE and then uploaded to a central server (e.g. a cloud host), and models are learned over the encrypted data by running machine learning algorithms adapted for AHE [Aono et al., 2016; Yuan and Yu, 2014]. In this way it could protect the raw data and achieve some kind of data privacy but doing arithmetic on encrypted data comes at a cost both in memory and processing time. And the data, although encrypted, cannot be kept locally, thus it raises potential risk of data leakage.

Instead of transmitting encrypted data to a central server, some intermediate results could be encrypted with AHE and transmitted during the training process. This brings some significant advantages: 1) the raw data are kept locally by each party, and 2) the amount of data to be encrypted is minimized through careful design, thus 3) the overall computation overhead could be reduced greatly. In this research direction, Hardy et al. [2017] proposed a solution for federated logistic regression on vertically partitioned data. Our work is another endeavor in this direction and we present a new solution for vertical federated logistic regression in this paper.

As compared with [Hardy et al., 2017], our solution is different in the following three perspectives. First, we adopt an architecture that removes the third-party coordinator, which can greatly simplify the system deployment. Second, we keep the loss function unchanged, instead of approximating it with a polynomial. Third, we aim to handle very large datasets and our solution is designed purposefully to be parallel distributed and scalable.

# 3 Solution

## 3.1 Overview of Logistic Regression

Logistic regression is a widely used machine learning model. In its basic form, it uses a logistic function to model a binary dependent variable, as described below:

$$P(y=1|x;\Theta) = h_\Theta(x) = \frac{1}{1+e^{-\Theta x}} \qquad (3)$$

In the setting of vertical federated learning of two parties $A$ and $B$, assuming two datasets $\{x_i^A | i \in D_A\}$, $\{x_i^B, y_i | i \in D_B, y_i \in \{0,1\}\}$, and model parameters $\Theta^A$, $\Theta^B$ corresponding to the feature space of $x^A$, $x^B$, respectively, the training objective is:

$$L = -\frac{1}{n}\sum_{i=1}^n y_i \log h_\Theta(x_i) + (1-y_i)\log(1-h_\Theta(x_i)) \qquad (4)$$

And the gradients are:

$$\frac{\partial L}{\partial \Theta} = -\frac{1}{n}\sum_{i=1}^n (y_i - h_\Theta(x_i))x_i \qquad (5)$$

And $\Theta x$, the linear predictor, can be written as:

$$\Theta x = \Theta^A x^A + \Theta^B x^B \qquad (6)$$

## 3.2 Protocol of Model Training

To train a federated logistic regression model, we need to protect both the raw data, i.e. $x_A$, $x_B$, and $y$, and model, i.e. $\Theta_A$ and $\Theta_B$, as summarized in Table 1. These data are stored locally by each party and are forbidden to be transmitted. It's also expected that these data will not be learned by others from any intermediate data transmitted during the model training.

|  | Data to be protected and stored locally |
|---|---|
| Party A | $x_A, \Theta_A, y$ |
| Party B | $x_B, \Theta_B$ |

Table 1: Data privacy

Table 2 shows the main steps of our proposed federated logistic regression training process. In practice, when training a model over a dataset, Step 2 to Step 6 will be usually iterated a lot of times until a maximum iteration number is reached or some convergence conditions are satisfied. Note that some intermediate data are exchanged but no data listed in Table 1 is transmitted between two parties during the model training. The major steps which involves intermediate data exchange that might lead to potential data leakage are listed below:

1) In Step 2, $B$ sends $\Theta_i^B x_i^B$ in plaintext to $A$ for each instance.
2) In Step 3, A sends $[\![(y_i - \hat{y}_i)]\!]$ to B.
3) In Step 4, B sends $\left[\!\left[\frac{\partial L}{\partial \Theta^B}\right]\!\right] + [\![R_B]\!]$ to A, where $[\![R_B]\!]$ are encrypted random masks to avoid A to learn the gradients of B.

The intermediate data exchanged during the model training are summarized in Table 3. We will further discuss the security of the model training protocol later in Section 3.4.

|  | Party A | Party B |
|---|---|---|
| Step 0 | Create an encryption key pair, and send the public key to $B$ | |
| Step 1 | Initialize $\Theta^A$ | Initialize $\Theta^B$ |
| Step 2 | Compute $\Theta^A x_i^A$ for $i \in D_A$ | Compute $\Theta^B x_i^B$ for $i \in D_B$ and send them to $A$ |
| Step 3 | Compute $\Theta x_i = \Theta^A x_i^A + \Theta^B x_i^B$, $\hat{y}_i = h_\Theta(x_i)$, $[\![(y_i - \hat{y}_i)]\!]$, and send $[\![(y_i - \hat{y}_i)]\!]$ to $B$ for $i \in D_A$ | |
| Step 4 | Compute $\frac{\partial L}{\partial \Theta^A}$ and the loss $L$ | Compute $\left[\!\left[\frac{\partial L}{\partial \Theta^B}\right]\!\right]$, generate random number $R_B$, and send $\left[\!\left[\frac{\partial L}{\partial \Theta^B}\right]\!\right] + [\![R_B]\!]$ to $A$ |
| Step 5 | Decrypt $\left[\!\left[\frac{\partial L}{\partial \Theta^B}\right]\!\right] + [\![R_B]\!]$, and send $\frac{\partial L}{\partial \Theta^B} + R_B$ to $B$ | |
| Step 6 | Update $\Theta^A$ | Update $\Theta^B$ |

Table 2: Model training protocol of logistic regression for vertical federated learning

|  | Intermediate results to be transmitted |
|---|---|
| Party A | $[\![y_i - \hat{y}_i]\!]$ for $i \in D_A$ |
| Party B | $\Theta^B x_i^B$ for $i \in D_A$, $\left[\!\left[\frac{\partial L}{\partial \Theta^B}\right]\!\right] + [\![R_B]\!]$ for each feature |

Table 3: Intermediate data exchange

### 3.3 System Architecture

We aim to build a parallel distributed system that can be deployed in real-world production systems to handle very large corpus of data very efficiently. Figure 3 shows the system architecture of our proposed solution. For each party, it adopts a worker-parameter server architecture, where there is a centralized parameter server and a set of worker nodes. There are no connections between worker nodes and the worker node can only communicate with its corresponding parameter server. The only communication channel between the party A and B exists between their corresponding parameter servers.

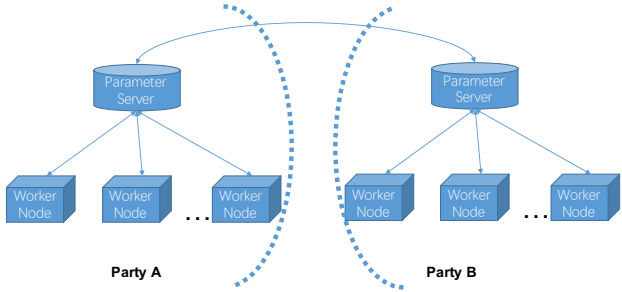

Figure 3: System architecture

The worker-parameter server architecture in our implementation is adapted from the open-sourced *ps-lite*[1] project. The communication between parameter servers is also built on *brpc*[2], a well-known open-sourced library for RPC. For secure intermediate results exchange, we utilize libpaillier[3] to implement additively homomorphic encryption. And most of codes are written in C++.

### 3.4 Security Analysis

In this Section, we analyze the security of the training protocol shown in Table 2.

**Security of A's label data.** Let's check whether B can learn $y_i$ for $i \in D_A$. Note that if B can learn the residuals $(y_i - \hat{y}_i)$ for $i \in D_A$, it can then learn the labels $y_i$ $i \in D_A$ because the residual will be negative if $y_i = 0$ and positive otherwise. We can find that in each iteration, only Step 3 and Step 5 involves transmitting data from A to B. In Step 3, the encrypted residuals $[\![(y_i - \hat{y}_i)]\!]$ for $i \in D_A$ are transmitted. Because the party B only has the public key, it cannot decrypt the encrypted residuals. In Step 5, the randomly masked gradients of B, $\frac{\partial L}{\partial \Theta^B} + R_B$, are transmitted, from which B can restore its gradients. From equation (5) we can find that the gradients are computed as a linear combination of residuals. Assuming that B has *n* instances and *m* features, we can get *m* linear combinations of *n* residuals. When $m \geq n$, these residuals can be learned by solving the systems of linear equations.

[1] https://github.com/dmlc/ps-lite
[2] https://github.com/apache/incubator-brpc
[3] http://acsc.cs.utexas.edu/libpaillier/

**Security of B's private data.** Let's check whether A can learn $x_i^B$ for $i \in D_B$. In each iteration, only Step 2 and Step 4 involves transmitting data from B to A. In Step 2, B sends $\Theta^B x_i^B$ for $i \in D_B$ to $A$. From these transmitted values A can construct *n* equations of $\Theta^B$ and $x_i^B$ for $i \in D_B$ where there are $(n+1) \times m$ unknowns. Note that both $\Theta^B$ and $x_i^B$ for $i \in D_B$ are unknown to A, and B has *n* instances and *m* features. It's obvious the system of equations cannot be solved because $n < (n+1) \times m$. If the steps are iterated one more time, A can get another *n* equations. Meanwhile, A also gets *m* new unknowns because $\Theta^B$ has been changed after each iteration. If the steps are iterated *r* rounds, A will get a total of $n \times r$ equations where there are $(n+r) \times m$ unknowns. When $n > m$ and $r \geq \frac{n \times m}{n-m}$, these unknowns can be learned by solving the system of equations.

In Step 4, B sends $\left[\!\left[\frac{\partial L}{\partial \Theta^B}\right]\!\right] + [\![R_B]\!]$ to A. A can construct *m* equations of $R_B$ and $x_i^B$ for $i \in D_B$ where there are $(n+1) \times m$ unknowns. It's obvious that the system of equations cannot be solved because $m < (n+1) \times m$. If the steps are iterated *r* rounds, A will get a total of $m \times r$ equations and there will be $(n+r) \times m$ unknowns. It's also obvious that the system of equations cannot be solved. Note that if the encrypted gradients are not masked with random numbers, A can learn $x_i^B$ for $i \in D_B$ after *n* rounds.

**Security of A's feature data.** Third, let's check whether B can learn $x_i^A$ for $i \in D_A$. As analyzed above, B can learn the residuals $(y_i - \hat{y}_i)$ for $i \in D_A$ in some circumstances and further learn $y_i$ for $i \in D_A$. As a result, B will also learn $\hat{y}_i$ for $i \in D_A$ and further learn $\Theta^A x_i^A$ for $i \in D_A$. Starting from here, we can apply the similar analysis logic discussed above to conclude that there are some circumstances where B can learn both $\Theta^A$ and $x_i^A$ for $i \in D_A$.

In the above analysis, we assume that both A and B are honest but curious. In the case that A is malicious, it can also learn $x_i^A$ one by one by setting all residuals to zero except one. To sum up, we can conclude that our solution is secure in practice through careful boundary condition checking.

## 4 Experiments

In this Section, we evaluate the performance of our solution through experiments from the perspectives of correctness, efficiency, and scalability.

### 4.1 Settings

**Datasets**. We evaluated the performance of our parallel distributed logistic regression for vertical federated learning over two datasets: Mnist[4] and Citeseer[5]. Table 2 describes some basic information about these two datasets. Note that Mnist is comprised of handwritten digits of 0 to 9 for multi-class classification. In the following experiments, we transform it to be suitable for binary classification: odd vs. even digits. Citeseer itself is a dataset for binary classification. It's very sparse and there are only 512,267 non-zero feature values. It's also highly skewed.

| Name | # Rows | # Cols | IsDense |
|---|---|---|---|
| Mnist | 70,000 | 784 | Yes |
| Citeseer | 181,395 | 105,354 | No |

Table 4: A brief description of datasets

**Environments**. The experiments are executed on Baidu cloud[6]. Each machine in our experiments has 16 CPU cores and 64 GB RAM. We allocate such a machine for each parameter server and each worker node. And in each experiment, one parameter server and multiple worker nodes are deployed for each party. The machines allocated for each party are hosted in the same region of Baidu cloud. The time-consuming computation is executed via multi-threading and the number of threads is set to 8 in the following experiments.

### 4.2 Results

In this Section we present the experimental results. Table 5 shows the AUC of learned modes on Mnist and Citeseer dataset. The convergence curve of AUC on Citeseer dataset is shown in Figure 4. And Figure 5 shows the scalability of the system and speedup on Citesser dataset with the increase of worker nodes. Note that because the Mnist dataset is relatively small and the training process is very fast even in single worker node, we did not perform the scalability experiment over the dataset.

The experimental results show that: 1) the learned models have excellent performance as measured by AUC; 2) the model quickly converges to a steady status after a number of iterations; and 3) the system is scalable and has sub-linear speedup. Besides, the overall running time is comparable with the traditional implementation of logistic regression. It proves that the system is suitable to be deployed in production system for handling vertical federated learning on very large corpus of decentralized data.

We also find some problems that might be the performance bottlenecks and we will fix these problems in near future. For example, we find that the training time can be further reduced if we offload some computations from parameter server to worker nodes.

|  | Train AUC | Test AUC |
|---|---|---|
| Mnist | 0.95 | 0.95 |
| Citeseer | 0.99 | 0.89 |

Table 5: AUC of learned models

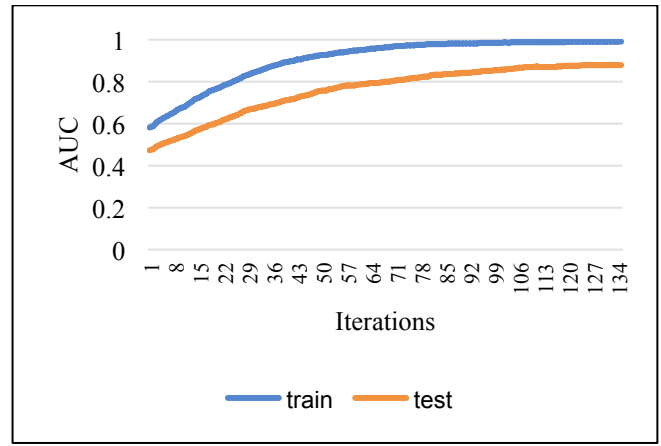

Figure 4: AUC Convergence on Citeseer dataset

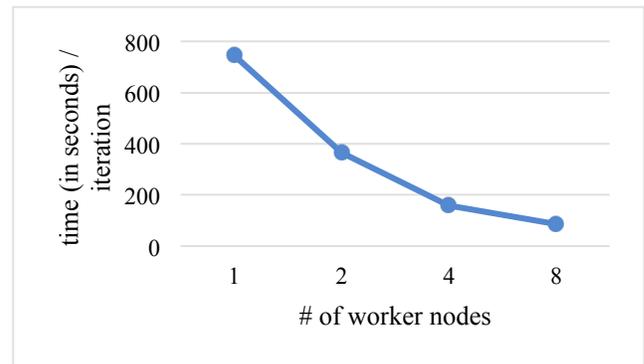

Figure 5: Scalability and speedup on Citeseer dataset

## 5 Conclusions

Federated learning is becoming a hotspot in machine learning community due to its ability to allow secure and efficient joint model training on a large corpus of decentralized data. In this paper, we present a solution for parallel distributed logistic regression for vertical federated learning. Dif-ferent with existing solutions, we remove the role of third-party coordinator from the system and allow any two parties to run federated logistic regression directly. This system design brings some significant advantages, e.g. avoiding the practical difficulty to find an authoritative third-party coordinator that can be trusted by all participants, and reducing the complexity of system deployment, etc. We also evaluated our implementation over two datasets (a sparse one and a dense one) and the experimental results show the great efficiency and scalability of the system. Moreover, our implementation can be easily extended to support the joint model training of multiple parities. In future, we will study the combination of logistic regression and deep neural network (DNN) for vertical federated learning. The basic idea is to learn hidden features from the input data with the help of DNN and feed them to the final logistic layer.

---

[4] http://yann.lecun.com/exdb/mnist/
[5] http://komarix.org/ac/ds/J_Lee.txt.bz2
[6] https://cloud.baidu.com/